\documentclass[conference]{IEEEtran}
\IEEEoverridecommandlockouts

\usepackage{booktabs} 
\usepackage[utf8]{inputenc} 
\usepackage{makecell} 
\usepackage{tabularx} 
\usepackage{algorithm} 
\usepackage{algpseudocode} 
\usepackage{csquotes} 
\usepackage{amsmath} 
\usepackage{subfigure}
\usepackage{multirow}
\usepackage{multicol}
\usepackage{graphicx, subfigure}
\usepackage{url}

\usepackage{tikz}
\usetikzlibrary{decorations.pathmorphing}

\def\BibTeX{{\rm B\kern-.05em{\sc i\kern-.025em b}\kern-.08em
    T\kern-.1667em\lower.7ex\hbox{E}\kern-.125emX}}

\makeatletter
\newcommand{\newlineauthors}{%
  \end{@IEEEauthorhalign}\hfill\mbox{}\par
  \mbox{}\hfill\begin{@IEEEauthorhalign}
}
\makeatother

\begin{document}
\title{Generation of Consistent Sets of Multi-Label Classification Rules with a Multi-Objective Evolutionary Algorithm}

\author{\IEEEauthorblockN{Thiago Zafalon Miranda}
\IEEEauthorblockA{\textit{Departamento de Computação} \\
\textit{Universidade Federal de São Carlos}\\
São Carlos, Brazil \\
thiago.zafalon.miranda@gmail.com}
\and
\IEEEauthorblockN{Diorge Brognara Sardinha}
\IEEEauthorblockA{\textit{Departamento de Computação} \\
\textit{Universidade Federal de São Carlos}\\
São Carlos, Brazil \\
diorgebs@gmail.com}
\and
\IEEEauthorblockN{Márcio Porto Basgalupp}
\IEEEauthorblockA{\textit{Instituto de Ciência e Tecnologia} \\
\textit{Universidade Federal de São Paulo}\\
São José dos Campos, Brazil \\
basgalupp@unifesp.br}\\
\newlineauthors
\IEEEauthorblockN{Yaochu Jin}
\IEEEauthorblockA{\textit{Department of Computer Science} \\
\textit{University of Surrey}\\
Guildford, United Kingdom \\
yaochu.jin@surrey.ac.uk}
\and
\IEEEauthorblockN{Ricardo Cerri}
\IEEEauthorblockA{\textit{Departamento de Computação} \\
\textit{Universidade Federal de São Carlos}\\
São Carlos, Brazil \\
cerri@ufscar.br}
}

\maketitle

\begin{abstract}
Multi-label classification consists in classifying an instance into two or more classes simultaneously. It is a very challenging task present in many real-world applications, such as classification of biology, image, video, audio, and text. Recently, the interest in interpretable classification models has grown, partially as a consequence of regulations such as the General Data Protection Regulation.
In this context, we propose a multi-objective evolutionary algorithm that generates multiple rule-based multi-label classification models, allowing users to choose among models that offer different compromises between predictive power and interpretability.
An important contribution of this work is that different from most algorithms,
which usually generate models based on lists (ordered collections) of rules, our algorithm generates models based on sets (unordered collections) of rules, increasing interpretability. Also, by employing a conflict avoidance algorithm during the rule-creation, every rule within a given model is guaranteed to be consistent with every other rule in the same model. Thus, no conflict resolution strategy is required, evolving simpler models. We conducted experiments on synthetic and real-world datasets and compared our results with state-of-the-art algorithms in terms of predictive performance (F-Score) and interpretability (model size), and demonstrate that our best models had comparable F-Score and smaller model sizes.
\end{abstract}



\section{Introduction}
\label{ref:intro}

Multi-label classification is a task in the machine learning field with applications in different areas,
such as  Bioinformatics~\cite{vens2008decision, guo2016human}, sentiment analysis~\cite{liu2015multi} and text classification~\cite{gonccalves2003preliminary, lauser2003automatic,luo2005evaluation}. It can be informally described as: given a collection of instances, each associated to a set of labels, and assuming that there exists a function $h$ that associates each object to its respective set of labels, try and generate a function $g$ that approximates the behavior of $h$.

In recent years, the interest in interpretable classification models has grown, partly due to regulations such as the General Data Protection Regulation (GDPR), a regulation \enquote{whereby users can ask for explanations of an algorithmic decision that significantly affects~them}~\cite{goodman2016european}.

It is generally accepted that rule-based classifiers are among the most interpretable models~\cite{freitas2014comprehensible}. The training phase of such classifiers usually consists of creating and tuning a list of classification rules. A classification rule usually has two components: an antecedent and a consequent. The antecedent is a collection of tests over feature values, and the consequent is the set of labels that will be assigned to the dataset instance if it passes all the antecedent's tests. A simple multi-label classification rule is exemplified as follows:

\begin{figure}[h]
\begin{center}
\begin{tabular}{l c c c c c}
IF      & 15        & $\leq$ & age              & < & 18 \\
AND     & 750       & $\leq$ & volume           & < & 1000 \\
AND     & 0         & $\leq$ & price            & < & 50 \\
THEN    & labels    & =      & \{aged, cheap\}    &   &\\
\end{tabular}
\end{center}
\end{figure}

To improve the interpretability of a classification model, researchers usually try to minimize the size of the model (e.g. the number of rules or tests in the rules' antecedents)~\cite{cerri2019inducing}. In~\cite{otero2013improving}, however, the authors suggest another approach: to employ sets (unordered collections) of rules, instead of lists (ordered collections). The premise is that in a list of rules, the n-th rule can not be correctly interpreted alone, because an instance it covers may also be covered by a previous rule; the actual classes predicted by the classifier would be the ones of the previous rule. On the other hand, a set of rules allows the user to analyze the rules individually, making the model more interpretable.
However, without the inherent \textit{order of application} imposed by a list, a collection of rules may become inconsistent if multiple rules (with different consequents) cover the same dataset instance. In~\cite{otero2013improving}, conflict resolution strategies are discussed. Such strategies allow the classifier to function correctly even if it contains multiple rules that contradict each other. In~\cite{miranda2019preventing}, the authors proposed a conflict avoidance algorithm, which supplements rule-creation processes to prevent such inconsistencies from arising, rendering conflict resolution strategies unnecessary.

In this work, we treat the generation and optimization of multi-label classification models considering both their predictive power and interpretability. Thus, we treat the classification task as a multi-objective optimization problem, and present a multi-objective evolutionary algorithm in which each individual of the population is a complete rule-based classification model, allowing the user to choose between models with different compromises between interpretability and predictive power. The models are based on sets of rules, and during the evolution the rules are generated using the conflict avoidance algorithm presented in~\cite{miranda2019preventing}, which guarantees that all rules within a given model are consistent\footnote{We present a formal definition for \enquote{consistency} in Section~\ref{sec:rule-creation}} with each~other.

The remainder of this paper is organized as follows. In Section~\ref{sec:related-works}, we briefly discuss related work. In Section~\ref{sec:minotaur}, we present the proposed algorithm. In Section~\ref{sec:exps}, we discuss the experiments we conducted and their results. Finally, in Section~\ref{sec:conclusion}, we present our final conclusions and future research directions.
\section{Related Work}
\label{sec:related-works}

In this Section,
we briefly discuss the literature about 
two important aspects of our work,
multi-label classification and interpretability.

\subsection{Multi-label Classification}
The two commonest approaches to multi-label classification are
problem transformation and
algorithm adaptation~\cite{tsoumakas2009mining}.
The former consists in transforming the multi-label problem into a
collection of single-label problems,
allowing the usage of traditional classification algorithms,
whose outputs are then combined to produce 
a multi-label prediction.
The algorithm adaptation approach consists in adapting single-label algorithms, or building new ones, to 
directly handle multi-label~problems.

One of the most intuitive problem transformation technique
is the Binary Relevance (BR)~\cite{zhang2018binary},
which consists in splitting the multi-label dataset into
$N$ binary datasets, 
with $N$ being the number of classes in the original dataset. A binary classifier is then trained for each dataset, and the individual binary predictions are aggregated to
produce the final multi-label prediction. The main drawback of this technique is that
it neglects possible relationships that may exist between classes. Also, as it requires training $N$ classifiers, the final model can be very complex in terms of interpretability.

A technique similar to BR is the
Classifier Chains (CC)~\cite{read2011classifier}.
It also trains multiple binary classifiers,
with the notable difference that the output of each 
classifier is concatenated with the instances' features,
which creates an augmented description of the instance. This augmented description is then provided to
the next classifier of the chain.
This augmentation process does manage to capture
some class relationships that are ignored in BR. However, it also means that the effectiveness of the chain
depends on the order of the classifiers in the chain.

Another recurrent problem transformation technique
is the Label Powerset~\cite{tsoumakas2007multi},
which transforms the multi-label dataset
into a single-label one by creating new classes
from the concatenation of sets of labels
present in the original dataset.
For instance, if there is an object whose labels
are $\{A, B, F\}$, a new class $[ABF]$ is created.
The technique,
unlike BR and CC,
does not suffer from the performance drawback
of training multiple classification models;
but it does tend to create too many classes that may end up with
too few positive instances associated with them.

One of the first work that employs the algorithm adaptation strategy
was proposed by Clare and King, 2001~\cite{clare2001knowledge},
which modifies decision trees
to perform hierarchical multi-label classification by
modifying their entropy function.
From the interpretability perspective,
the work is also interesting,
because after the tree model is created,
the authors \enquote{break} the rules that predict
multiple labels into multiple rules that predict
single labels.

In~\cite{zhang2005k},
the authors propose the first lazy multi-label algorithm;
an adaptation of k-Nearest Neighbors.
The proposed algorithm considers the frequency of classes
in the neighborhood of the instance being classified to 
generate the multi-label prediction.

To the best of author's knowledge, Cerri et al., 2019~\cite{cerri2019inducing} proposed the first genetic algorithm that generates a single global model
for hierarchical multi-label classification.
In the proposed algorithm, named HMC-GA,
each individual represents a single classification rule,
and the entire population represents the classification model.

In~\cite{vens2008decision},
the authors present the Clus-HMC algorithm,
which generates predictive clustering trees.
The algorithm views the tree as a hierarchy of clusters,
maximizing the similarities of instances in each cluster.
The implementation provided by the authors contains a parameter
to generate a set of rules from the tree (each path
from the root to a leaf becomes a rule).

\subsection{Interpretability}
In the context of classification models,
interpretability is not an objectively and consistently
defined concept~\cite{lipton2016mythos};
different authors use the term to refer
to different characteristics a model can have.
However, it is reasonable to say that some types of
classification models are inherently more interpretable
than others; e.g.:
a decision tree~\cite{quinlan1986induction}
can be said to be more interpretable than 
a deep neural network~\cite{lecun2015deep}.

Rule-based classifiers are considered to be among
the most interpretable models~\cite{freitas2014comprehensible}.
For such classification models, 
the number of rules is often used to quantify the interpretability of the models\cite{cerri2012genetic, otero2013improving}.

In~\cite{otero2013improving},
the authors discuss another way of improving
the interpretability of such models:
to employ sets (unordered collections) of rules,
instead of lists (ordered collections) of rules.
Indeed,
if the collection of rules has no intrinsic ordering,
then one could, in principle,
analyze a particular rule in isolation.

Frequently, however,
removing the ordering of the collection may render
the classification inconsistent,
for it may contain rules that cover the same feature-space region,
but assign different labels to this region.
In such cases, 
it becomes necessary to utilize a mechanism
to decide which rule should be used.
In~\cite{otero2013improving},
the authors described such mechanisms as
\enquote{conflict resolution strategies}.
An example of such mechanism is:
whenever a conflict between two rules arises,
use the one that has the highest F-Score in the training~dataset.

In~\cite{miranda2019preventing},
the authors discuss another solution to the consistency
issue of sets of rules;
a conflict avoidance algorithm,
named Constrained Feature-Space Box-Enlargement (CFSBE),
which can find feature-space regions not covered
by any existing rules.
Creating a rule inside such a region
(i.e. a rule whose antecedent only covers such a sub-region),
ensures that the new rule will be consistent with the already existing rules.

In the next section, we show how CFSBE is used in the proposed evolutionary algorithm to induce consistent sets of multi-label classification rules.

\section{Proposed Algorithm}
\label{sec:minotaur}

\sloppy
Our algorithm, named
\textit{\textbf{M}ulti-object\textbf{i}ve evolutio\textbf{n}ary alg\textbf{o}rithm for consis\textbf{t}ent and interpret\textbf{a}ble m\textbf{u}lti-label \textbf{r}ules} (MINOTAUR)
is a multi-objective evolutionary algorithm
that generates rule-based classification models. Each individual of the population is a non-empty set of multi-label classification rules,
that is, a complete classification model.
Each model may contain an arbitrarily large number of rules,
as long as the rules of the model are consistent with each other.

A set of rules (a classification model) is said to be consistent
if all its rules are consistent with each other. A pair of rules is said to be consistent 
if their consequents are equal or 
if the feature-space regions described by their antecedents
does not overlap~\cite{miranda2019preventing}.

In our proposal, the antecedent of a rule is a set of \textit{feature-tests},
and it must contain exactly one \textit{feature-test}
for each feature of the dataset.
A \textit{feature-test} is a function that checks whether 
the value of a given feature is inside an interval.
An example of a \textit{feature-test} would be $ft(age) = 0 \leq age < 18$.

It is important that one of the comparisons of the \textit{feature-test} is inclusive and the other exclusive,
so that feature-tests of different rules can \enquote{touch} each other, that is,
the lower bound of one rule \textit{feature-test} may be equal to
the upper bound of another rule \textit{feature-test}.
If both tests were inclusive,
rules that contain feature-tests that \enquote{touch} each other would be inconsistent.
The rules $r_1$ and $r_2$ below exemplify this concept when applied to an instance with an attribute value $w = 50$.
It is possible to see that, in this case, the rules are inconsistent, i.e., both rules cover the instance.
\begin{figure}[h]
\begin{center}
\begin{tabularx}{0.91\columnwidth}{Xcccccccl}
$r_1 :=$ & IF & 0   & $\leq$ & w & $\leq$ & 50       & THEN & class=light \\
$r_2 :=$ & IF & 50  & $\leq$ & w & $\leq$ & $\infty$ & THEN & class=heavy \\
\end{tabularx}
\end{center}
\end{figure}

The lower and upper bounds of a \textit{feature-test} can assume any value present in
the dataset for that feature in addition to $-\infty$ and $\infty$.
\textit{Feature-tests} with infinity as boundaries are tautological and indicate
that a given feature is not important for that rule,
and that such \textit{features-test} may be removed after MINOTAUR finishes running
(effectively reducing the size of the rules).
During the evolution, however, they must be preserved,
since CFSBE requires that all rules have the same size.
An example of such a \textit{feature-test} is:
$ft(age) = -\infty \leq age < \infty$.
It is also possible that only one of the comparisons
of the \textit{feature-test} becomes tautological,
which indicates that the \textit{feature-test}
can not be removed, but can at least be simplified.

The consequent of a rule is a binary vector.
If the n-th position of the vector is 1,
it indicates that the dataset instance belongs to the n-th class.
If it is 0, it indicates that the dataset instance does not belong
to the n-th class.

\begin{figure}[htbp]
\centering
\includegraphics[scale=0.55]{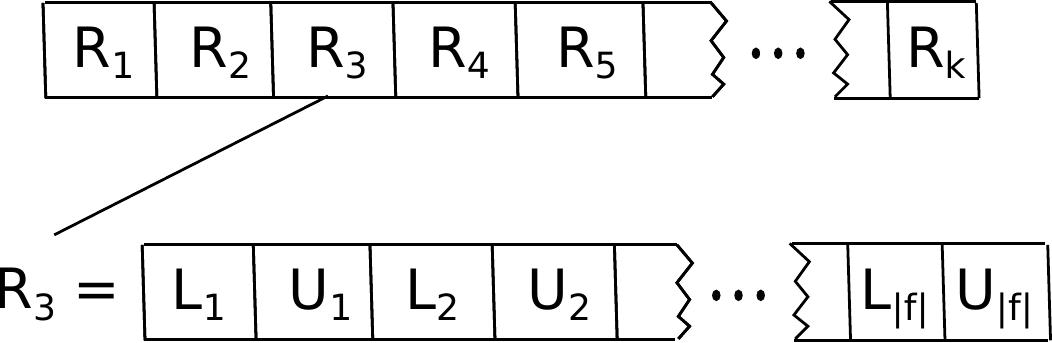}
\caption{Individual representation}
\label{fig:cerri-rules}
\end{figure}

A representation of our proposed individual is shown
in Figure~\ref{fig:cerri-rules}.
In this representation,
an individual is a vector of $k$ rules,
and each rule is a vector of size $2 \times |f|$, with $f$ the set of features of the dataset. Every two positions of a rule $R$ forms a tuple $(L_i, U_i)$,
each representing the lower bound and the upper bound
of the i-th \textit{feature-test} of such rule. Thus, every test in our rules verifies if a given attribute value is greater than or equal to a lower bound value, and if it is less than an upper bound value ($L_i \leq w < U_i$). 

\begin{figure}[htbp]
\centering
\includegraphics[scale=0.7]{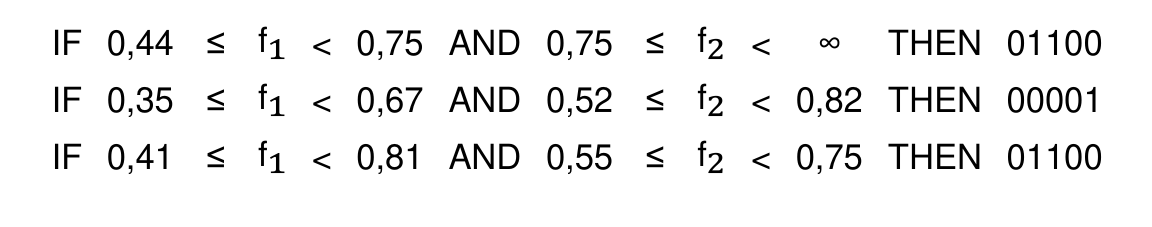}
\caption{Example of classification model}
\label{fig:model}
\end{figure}

To help visualize what a classification model is,
we present in Figure~\ref{fig:model} an actual classification
model generated for a dataset with two features and five classes. The prediction of a given model for a given dataset instance is
the consequent of the rule that matches the instance.
If no rule matches the instance,
we utilize a fallback mechanism commonly called \enquote{default rule}.
Many algorithms besides MINOTAUR, such as HMC-GA and Clus-HMC, employ this mechanism,
which assigns the dataset average set of labels (the average of the binary vectors) to the instance being classified. 
Since this \enquote{rule} does not contain an antecedent,
and is only activated under special circumstances,
it is not counted towards the model size for any algorithm.

The MINOTAUR algorithm contains two steps:
initialization, in which the initial population is created,
and the mutation-selection loop,
in which new individuals are created and the fittest are selected.
The mutation-selection loop stops when at least one of the following conditions is met:
the maximum number of generations is reached,
or the maximum number of failed mutation attempts per generation is reached 
(see Section~\ref{sec:mutation-selection}). 
We have decided to disable crossover operators since they have not contributed improving the results according to preliminary experiments. In Section~\ref{sec:rule-creation},
we describe the rule creation mechanism
and the initialization phase.
In Section~\ref{sec:mutation-selection},
we describe the mutation-selection loop.

\subsection{Rule Creation and Initialization}
\label{sec:rule-creation}

A centerpiece of MINOTAUR is the rule creation mechanism,
employed both during the initialization phase
and by certain mutation operators.
The rule creation mechanism itself is built on top
of the CFSBE algorithm~\cite{miranda2019preventing}, which we will summarize below.

CFSBE has three input parameters.
The first, named $R$, is a (possibly empty) set of consistent rules.
The second input parameter, named $seed$,
is a dataset instance that is not covered by any rule
of $R$. 
The last input parameter, named $O$,
is a permutation of $\{1, 2, \ldots, |f|\}$,
with $f$ being the set of features of the dataset.

The output of CFSBE is a $|f|$ dimensional hyperrectangle,
named $box$.
This $box$ contains the $seed$,
do not overlap with the feature-space regions described by the rules of $R$, and cannot be further expanded.
To generate such a $box$,
CFSBE creates a hyperrectangle around the $seed$
and enlarges it along each dimension until a further expansion would result in an overlap between the $box$ and the feature-space regions described by the rules of $R$.
If $R$ is empty, then the $box$ covers the entire feature-space.

The parameter $O$ indicates the order in which the dimensions should be
expanded.
For example, if $O=(3,1,2)$, then the $box$ would be enlarged
first along the third dimension,
then along the first dimension,
and finally along the second dimension.

\begin{figure}[htbp]
       \center
       \subfigure[refa][Enlarging $f_1$ first]{\includegraphics[scale=0.2]{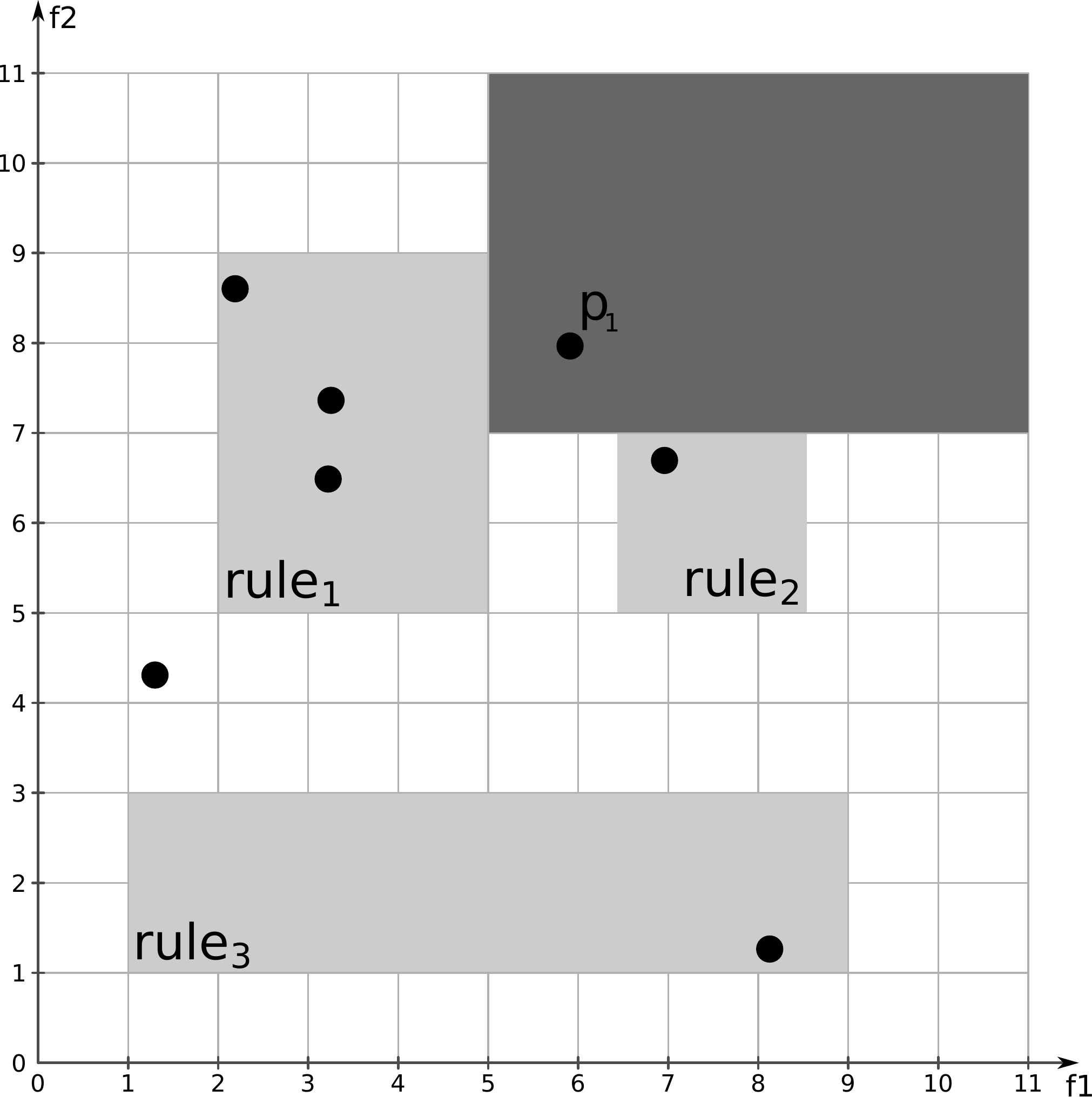}}
       \hspace{0.3em}
       \subfigure[refb][Enlarging $f_2$ first]{\includegraphics[scale=0.2]{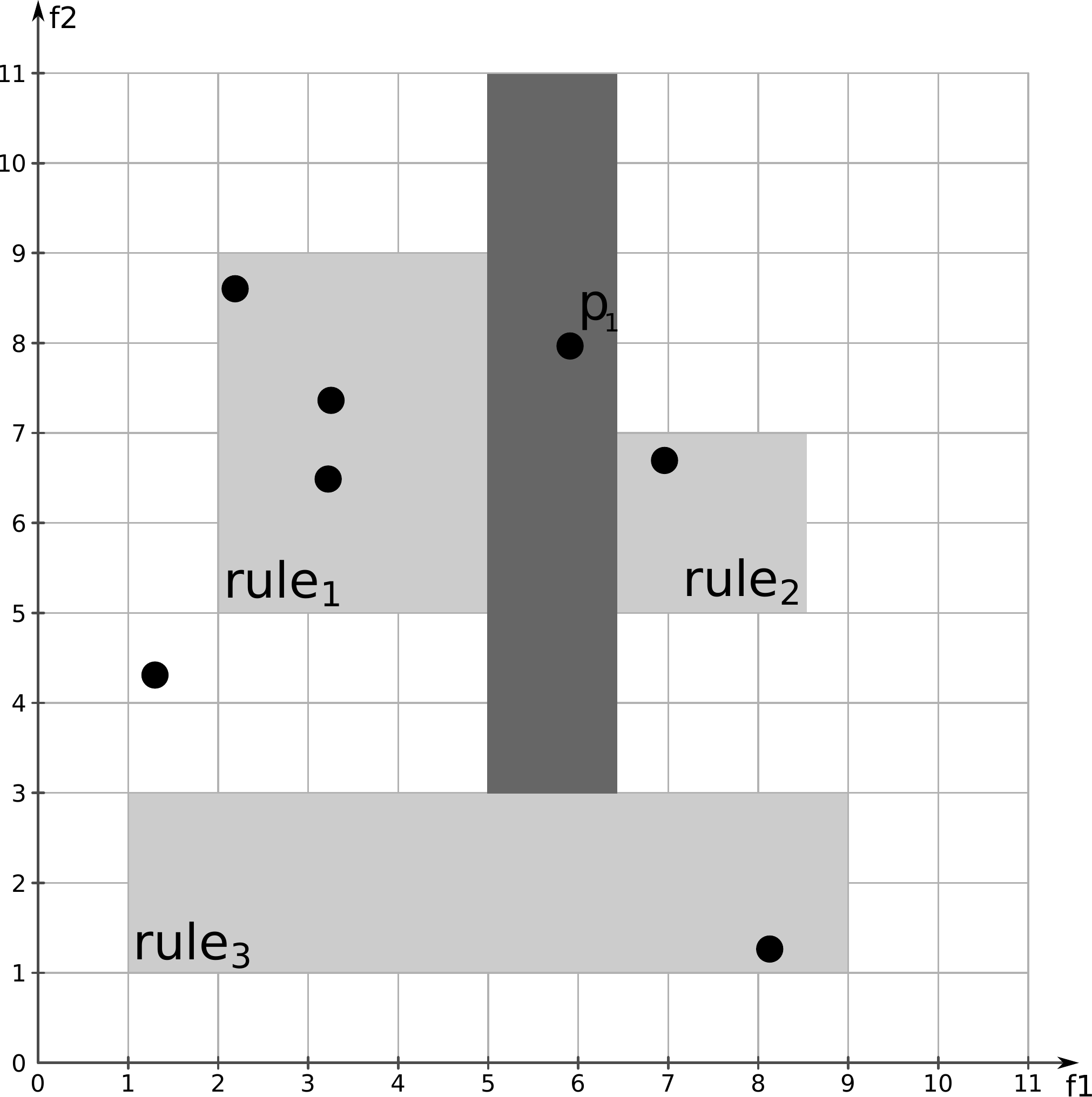}}
       \caption{Box enlargment ilustration}
       \label{fig:box-enlarg}
\end{figure}

To illustrate how different values of $O$ generate
different $boxes$,
consider Figure~\ref{fig:box-enlarg} (adapted from~\cite{miranda2019preventing}),
in which black dots represent dataset instances,
the dark-grey rectangle represents the output $box$,
the remaining rectangles represent the rules of $R$,
and $p_1$ is the seed.
Figure~\ref{fig:box-enlarg}(a) shows the resulting $box$ when $O=(1,2)$, and Figure~\ref{fig:box-enlarg}(b)
shows the resulting $box$ when $O=(2,1)$.

There are many ways to generate a rule from a $box$;
for instance, one could,
for each dimension of the $box$,
utilize the boundary values of the $box$ in that dimension
as the boundaries of a \textit{feature-test}.
This approach is not ideal because if $R=\emptyset$
the generated $box$, and the generated rule-antecedent,
would cover the entire feature-space,
preventing the creation of more rules.

In our algorithm, we adopted the following approach:
if we are trying to generate a new rule for a individual
$ind$, then we run CFSBE with $R=ind.rules$,
where $ind.rules$ is the current set of rules of individual $ind$,
a randomly generated $O$, and a $seed$ randomly chosen among
the possible dataset instances (i.e. the dataset instances not covered
by the rules of $ind$).
We then create a new box around the $seed$,
called $box_2$,
and enlarge it to try to cover $t$ (a hyperparameter of
MINOTAUR) dataset instances that are closest to the seed
and also not covered by the rules of the input set $R$.
The enlargement process stops when
the $box_2$ boundaries reach the boundaries of the CFSBE $box$
or when $t$ instances are covered.
We then create \textit{feature-tests} from the $box_2$ boundaries;
and such \textit{feature-tests} form the antecedent of the rule.
To create the consequent of the rule,
we average the labels of the dataset instances covered by the antecedent.

The mechanism described above is also used in the initialization phase of our proposal
to create the individuals of the first generation.
Such individuals contain a single rule.
To generate the first rule of each individual,
we use the method we just described,
but now using $R = \emptyset$.
It is important to observe that although we are using
the same value of $R$ for the creation of all first-rules,
such rules are not necessarily equal,
since the values of $seed$ and $O$ are random.

\subsection{Mutation and Selection}
\label{sec:mutation-selection}
Our mutation phase consists of a reproduction strategy which tries to create $m$ new individuals by cloning existing ones and applying specific mutation operators in the rules of these clones.
To do so, we randomly select an individual from the population, a mutation operator, 
and check if it is possible to apply such mutation to the selected individual.
If it is possible,
we create a clone of the individual and mutate it.
If it is not possible,
we increment a \textit{failed mutation attempts} counter
and randomly select another (individual, mutation-operator) pair.
To exemplify a mutation failure,
consider an individual $ind$ whose rules already cover all dataset instances.
Consider now that we want to create a new individual by cloning $ind$ and applying a mutation operator that adds a new rule to the clone.
Since the rules from $ind$ already cover all instances,
there is no suitable $seed$ for the execution of CFSBE. It is important to observe that our reproduction strategy consists of mutating a clone.
Thus, the original individual is not modified.
After all mutants are generated,
they are added to the population, and both the original individuals and their mutated-clones compete during the selection phase, which is performed using the NSGA-II algorithm~\cite{deb2002fast}.

We implemented three mutation operators,
the first is \textit{adding a new rule} to the selected individual.
The new rule is created using the process described in Section~\ref{sec:rule-creation},
with $R$ as the set of rules of the individual being mutated,
$O$ randomly generated and $seed$ randomly chosen among the possible
dataset instances.
The mutation can fail if $R$ already covers the entire feature-space
(or at least all the dataset instances).

The second type of operator is \textit{removing a rule},
which is done by randomly selecting a rule from the individual
and removing it.
More specifically, 
we shuffle the individuals' vector of rules and drop the last one.
This mutation can fail if the individual contains a single rule,
because removing it would effectively destroy the~model.

The last operator is \textit{substituting a rule},
which is done by randomly selecting a rule from the individual,
removing this rule, and creating a new one using the 
mechanism described in Section~\ref{sec:rule-creation}.
To create this new rule,
we run CFSBE with $R$ as the reduced set of rules
(i.e. the individuals' rules minus the removed one),
a randomly generated $O$, and a $seed$ randomly chosen
among the possible dataset instances
(i.e. the ones not covered by the reduced set of rules).
The newly generated rule is then added to the individual.
This mutation never fails, because if a rule is removed,
then it is always possible to create a new rule,
even if such rule only covers a sub-region of the previous one.

Each mutation operator has a weight associated with it,
which makes some types of mutations more likely to occur.
The weight for \textit{adding a new rule} is 1,
for \textit{removing a rule} is 2
and for \textit{substituting a rule} is 4.
To decide which mutation will occur,
we generate a random number between 1 and 7;
if the value is 7, we add a new rule;
if the value is 6 or 5, we remove a rule;
else we substitute a rule.
We chose such values to decrease the likelihood of a mutation failing,
which wastes computational time.
Further experimentation with different values can be conducted in future works.

After the desired number of mutants $m$ is generated,
the \textit{failed mutation attempts counter} is reset,
the mutants are added to the population,
and the fittest individuals of the population are selected using NSGA-II,
using as fitness a tuple with the micro-averaged F-Score~\cite{tsoumakas2009mining}
and the model size (number of rules) of the individual.
If, howeverm the \textit{failed mutation attempts counter}
reaches a certain threshold (a hyperparameter of the algorithm),
before $m$ new individuals were generated,
the evolution stops.
\section{Experiments}
\label{sec:exps}

To assess the effectiveness of our algorithm,
we conducted 10-fold cross-validated experiments with eight datasets.
Four of them are real-world ones publicly available\footnote{\url{http://mulan.sourceforge.net/datasets-mlc.html}}: 
CAL500~\cite{turnbull2008semantic},
emotions~\cite{trohidis2008multi},
scene~\cite{boutell2004learning},
and yeast~\cite{ueda2003parametric}. 
The other ones were generated with the Read et al., 2012~\cite{read2012scalable} multi-label proposal with four generators:
hyper-plane (synthetic0),
radial basis function (synthetic1),
random tree (synthetic2),
and wave-form (synthetic3). Table~\ref{tab:dataset-descriptions} shows the number of features, classes and instances of each dataset.

\begin{table}[tp]
\centering
\caption{Dataset Descriptions}
\begin{tabularx}{0.8\columnwidth}{Xccc}
\toprule
Name        & \#Features   & \#Classes    & \#Instances \\
\midrule
CAL500      & 68            & 174           & 502 \\
emotions    & 72            & 6             & 593 \\
scene       & 294           & 6             & 2407 \\
synthetic0  & 10            & 5             & 10000 \\
synthetic1  & 80            & 22            & 10000 \\
synthetic2  & 30            & 8             & 10000 \\
synthetic3  & 21            & 7             & 10000 \\
yeast       & 103           & 14            & 2417 \\
\bottomrule
\end{tabularx}
\label{tab:dataset-descriptions}
\end{table}

We compared the micro-averaged F-Score
of the highest scoring model generated by MINOTAUR
(remember that it generates a collection of models)
and the models generated by known single-label algorithms:
J48,
Support Vector Machines (SVM),
Naive Bayes (NB)
and k-Nearest Neighbors (k-NN).
In order to use such algorithms in a multi-label context
we used the problem transformation methods
classifier chains (CC), 
Binary Relevance (BR)
and Label Powerset (LP).
We have also conducted experiments with
the state-of-the-art multi-label algorithms HMC-GA~\cite{cerri2019inducing}
and Clus-HMC~\cite{blockeel1998top}. While HMC-GA is a genetic algorithm specifically proposed to generate lists of multi-label rules, Clus-HMC generates a single multi-label decision tree, which can then be converted into a set of~rules.

Since MINOTAUR and HMC-GA are non-deterministic methods, we executed them 30 times in each fold. For the other methods, only one execution per fold was necessary since they are deterministic.
Because MINOTAUR generates a collection of models, instead of a single one,
we selected its best model (i.e. with highest F-Score) in the training data,
and show its results obtained in the test partition.
We consider the problem transformation methods as the baselines focusing only on predictive performance,
while Clus-HMC and HMC-GA are baselines for a compromise between performance and interpretability.
We aim at showing that MINOTAUR has competitive performance in comparison with all methods while obtaining more interpretable models, i.e, models with fewer~rules.

Table~\ref{tab:fscores} presents the micro-averaged F-Scores obtained from all methods investigated. 
The best results are highlighted in bold face.
The standard deviations are presented only for MINOTAUR and HMC-GA,
and are obtained averaging the 30 executions in each fold.
This was performed to analyze the stability of the algorithms.

The small standard deviations in MINOTAUR's F-Scores suggest
that it is indeed stable and the F-Scores themselves indicate that our algorithm
is overall comparable with multi-label rule-based algorithms from the literature. It is interesting to observe that the models generated with problem-adaptation strategies,
although theoretically unable to capture as many relations between classes as the multi-label algorithms, did outperform them.
However, 
it is reasonable to assume that this is more likely a consequence of the 
performance-oriented approach employed, that sacrifices interpretability,
than a multi-label / single-label issue.
Consider, for instance, the J48 algorithm,
which generates an interpretable model (a classification tree).
One could argue that when the trees are combined to form a Classifier Chain,
or their predictions are combined to in Binary Relevance,
their interpretability is greatly reduced.
If we now consider LP-J48, that generates a single tree,
preserving the interpretability of the model,
we can see that its F-Scores are overall not very different from the ones obtained by the multi-label rule-based methods.


\begin{table*}[htbp]
\centering
\setlength{\tabcolsep}{4.3pt}
\caption{Micro-averaged F-Scores. Since MINOTAUR and HMC-GA are non-deterministic methods, we also present the standard-deviation ($\pm$) over 30 runs}
\begin{tabular}{lccccccccccccccc}
\toprule
\multirow{2}{*}{Dataset} & \multicolumn{3}{c}{Multi-label Rule-based Methods} & \multicolumn{4}{c}{Classifier Chains} & \multicolumn{4}{c}{Binary Relevance} & \multicolumn{4}{c}{Label Powerset} \\
& MINOTAUR & HMC-GA & Clus-HMC & J48  & SVM  & NB   & k-NN & J48  & SVM  & NB   & k-NN & J48  & SVM  & NB   & k-NN\\
\midrule
    CAL500 & 0.31 $\pm$ 0.00  & 0.34 $\pm$ 0.01  & 0.31  & {\bf 0.36}  & 0.34  & 0.29  & 0.34  & 0.35  & 0.33  & 0.33  & 0.34  & 0.33  & 0.34  & 0.34  & 0.34 \\
    emotions & 0.36  $\pm$ 0.02 & 0.49 $\pm$ 0.03  & 0.59  & 0.60  & 0.68  & 0.66  & 0.62  & 0.60  & 0.65  & 0.66  & 0.62  & 0.59  & {\bf 0.70}  & 0.62  & 0.62 \\
    scene & 0.31 $\pm$ 0.01  & 0.45 $\pm$ 0.03  & 0.60  & 0.63  & 0.70  & 0.56  & 0.69  & 0.62  & 0.68  & 0.56  & 0.69  & 0.59  & {\bf 0.74}  & 0.64  & 0.69 \\
    synthetic0 & 0.59 $\pm$ 0.00  & 0.48 $\pm$ 0.01  & 0.52  & 0.58  & 0.60  & 0.60  & 0.49  & 0.56  & 0.56  & 0.58  & 0.49  & 0.49  & {\bf 0.61}  & 0.61  & 0.49 \\
    synthetic1 & 0.02 $\pm$ 0.00  & 0.24 $\pm$ 0.01  & 0.12  & 0.39  & 0.15  & 0.08  & {\bf 0.65}  & 0.47  & 0.03  & 0.03  & {\bf 0.65}  & 0.26  & 0.24  & 0.13  & {\bf 0.65} \\
    synthetic2 & 0.47 $\pm$ 0.00  & 0.38 $\pm$ 0.01  & 0.34  & 0.50  & 0.48  & {\bf 0.51}  & 0.41  & 0.37  & 0.41  & 0.39  & 0.41  & 0.44  & 0.50  & 0.50  & 0.41 \\
    synthetic3 & 0.64 $\pm$ 0.00  & 0.61 $\pm$ 0.01  & 0.66  & 0.67  & 0.67  & 0.67  & 0.62  & 0.68  & 0.65  & 0.65  & 0.62  & 0.65  & 0.71  & {\bf 0.69}  & 0.62 \\
    yeast & 0.55 $\pm$ 0.00  & 0.56 $\pm$ 0.01  & 0.56  & 0.55  & {\bf 0.64}  & 0.54  & 0.60  & 0.58  & 0.63  & 0.55  & 0.60  & 0.54  & 0.64  & 0.60  & 0.60 \\
    \midrule
     rank  & 11.56 & 11.75 & 11.06 & 6.38  & 4.25  & 8.38  & 7.5   & 7.38  & 8.19  & 10    & 7.5   & 10.44 & \textbf{2.81} & 5.31  & 7.5 \\
\bottomrule
\end{tabular}
\label{tab:fscores}
\end{table*}

In Table~\ref{tab:mlc-interp},
we show the average (across the folds) number of rules in each model, highlighting the best results in bold face.
It is interesting to note that Clus-HMC's model for CAL500 indeed contains no rules.
This is because Clus-HMC applies a pruning procedure to the generated tree, which is then converted into rules. The pre-pruning model for CAL500 was not empty, but after pruning all rules were removed and only the fallback mechanism (the \enquote{default rule}) remained.
We also note that in this specific case of CAL500, the \enquote{default rule} obtained a competitive performance, although slightly inferior, with the other methods investigated. This suggests that CAL500 is a very challenging dataset, which can indeed be confirmed by its large number of classes (174) and small number of instances (502) when compared to the other datasets.
In six of the remaining seven datasets, MINOTAUR generated the smallest models,
while remaining competitive.
In the yeast dataset, for instance, all three algorithms had comparable
F-Scores, but MINOTAUR was able to achieve that with only 16 rules,
while  HMC-GA and Clus-HMC required 27 and 98 rules, respectively.



\begin{table}[htbp]
\centering 
\setlength{\tabcolsep}{8.5pt}
\caption{Number of Rules of the Multi-label Methods}
\begin{tabular}{llll}
\toprule
Dataset & \multicolumn{1}{c}{MINOTAUR} & \multicolumn{1}{c}{HMC-GA}  & \multicolumn{1}{c}{Clus-HMC}  \\
\midrule
    CAL500 & 27.94 $\pm$ 1.43  & 9.65  $\pm$  1.18  & {\bf 0.00} \\
    emotions & 11.03  $\pm$ 1.63  & \textbf{8.86  $\pm$  0.96}  & 21.30 \\
    scene & {\bf 10.41  $\pm$  0.98}  & 17.11 $\pm$  1.33  & 67.90 \\
    synthetic0 & {\bf 9.26  $\pm$  1.17}  & 37.31 $\pm$  4.19  & 13.70 \\
    synthetic1 & {\bf 12.36 $\pm$  0.75}   & 399.42 $\pm$  8.22 & 103.50 \\
    synthetic2 & {\bf 11.25  $\pm$  1.96}  & 145.76 $\pm$  3.54  & 40.90 \\
    synthetic3 & {\bf 13.38 $\pm$ 1.54}  & 66.54 $\pm$  2.82  &  32.90\\
    yeast & {\bf 15.93 $\pm$ 2.07}  & 26.74 $\pm$  2.32  & 97.90 \\
    \midrule
    rank  & \multicolumn{1}{l}{\bf 1.38} & \multicolumn{1}{l}{2.38}  & \multicolumn{1}{l}{2.25}  \\
\bottomrule
\end{tabular}
\label{tab:mlc-interp}
\end{table}

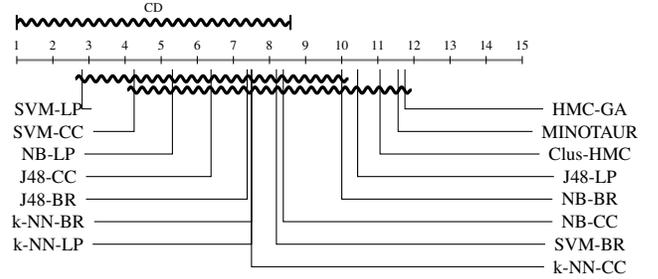
\begin{figure}[htbp] \centering \begin{tikzpicture}[xscale=1.8]
\node (Label) at  (01.2773,0.7) {\tiny{CD}}; 
\draw[decorate,decoration={snake,amplitude=.4mm,segment length=1.5mm,post length=0mm}, very thick, color = black](00.2667, 0.5) -- (02.2880, 0.5);
\foreach \x in {00.2667,02.2880} \draw[thick,color = black] (\x, 0.4) -- (\x, 0.6);

\draw[gray, thick](00.2667, 0) -- (04.0000, 0);
\foreach \x in {00.2667,00.5333,00.8000,01.0667,01.3333,01.6000,01.8667,02.1333,02.4000,02.6667,02.9333,03.2000,03.4667,03.7333,04.0000}\draw (\x cm,1.5pt) -- (\x cm, -1.5pt);
\node (Label) at (00.2667,0.2) {\tiny{1}};
\node (Label) at (00.5333,0.2) {\tiny{2}};
\node (Label) at (00.8000,0.2) {\tiny{3}};
\node (Label) at (01.0667,0.2) {\tiny{4}};
\node (Label) at (01.3333,0.2) {\tiny{5}};
\node (Label) at (01.6000,0.2) {\tiny{6}};
\node (Label) at (01.8667,0.2) {\tiny{7}};
\node (Label) at (02.1333,0.2) {\tiny{8}};
\node (Label) at (02.4000,0.2) {\tiny{9}};
\node (Label) at (02.6667,0.2) {\tiny{10}};
\node (Label) at (02.9333,0.2) {\tiny{11}};
\node (Label) at (03.2000,0.2) {\tiny{12}};
\node (Label) at (03.4667,0.2) {\tiny{13}};
\node (Label) at (03.7333,0.2) {\tiny{14}};
\node (Label) at (04.0000,0.2) {\tiny{15}};
\draw[decorate,decoration={snake,amplitude=.4mm,segment length=1.5mm,post length=0mm}, very thick, color = black](00.6993,-00.2500) -- ( 02.7167,-00.2500);
\draw[decorate,decoration={snake,amplitude=.4mm,segment length=1.5mm,post length=0mm}, very thick, color = black](01.0833,-00.4000) -- ( 03.1833,-00.4000);
\node (Point) at (00.7493, 0){};  \node (Label) at (0.5,-00.6500){\scriptsize{SVM-LP}}; \draw (Point) |- (Label);
\node (Point) at (01.1333, 0){};  \node (Label) at (0.5,-00.9500){\scriptsize{SVM-CC}}; \draw (Point) |- (Label);
\node (Point) at (01.4160, 0){};  \node (Label) at (0.5,-01.2500){\scriptsize{NB-LP}}; \draw (Point) |- (Label);
\node (Point) at (01.7013, 0){};  \node (Label) at (0.5,-01.5500){\scriptsize{J48-CC}}; \draw (Point) |- (Label);
\node (Point) at (01.9680, 0){};  \node (Label) at (0.5,-01.8500){\scriptsize{J48-BR}}; \draw (Point) |- (Label);
\node (Point) at (02.0000, 0){};  \node (Label) at (0.5,-02.1500){\scriptsize{k-NN-BR}}; \draw (Point) |- (Label);
\node (Point) at (02.0000, 0){};  \node (Label) at (0.5,-02.4500){\scriptsize{k-NN-LP}}; \draw (Point) |- (Label);
\node (Point) at (03.1333, 0){};  \node (Label) at (4.5,-00.6500){\scriptsize{HMC-GA}}; \draw (Point) |- (Label);
\node (Point) at (03.0827, 0){};  \node (Label) at (4.5,-00.9500){\scriptsize{MINOTAUR}}; \draw (Point) |- (Label);
\node (Point) at (02.9493, 0){};  \node (Label) at (4.5,-01.2500){\scriptsize{Clus-HMC}}; \draw (Point) |- (Label);
\node (Point) at (02.7840, 0){};  \node (Label) at (4.5,-01.5500){\scriptsize{J48-LP}}; \draw (Point) |- (Label);
\node (Point) at (02.6667, 0){};  \node (Label) at (4.5,-01.8500){\scriptsize{NB-BR}}; \draw (Point) |- (Label);
\node (Point) at (02.2347, 0){};  \node (Label) at (4.5,-02.1500){\scriptsize{NB-CC}}; \draw (Point) |- (Label);
\node (Point) at (02.1840, 0){};  \node (Label) at (4.5,-02.4500){\scriptsize{SVM-BR}}; \draw (Point) |- (Label);
\node (Point) at (02.0000, 0){};  \node (Label) at (4.5,-02.7500){\scriptsize{k-NN-CC}}; \draw (Point) |- (Label);
\end{tikzpicture}
\caption{Critical diagrams showing average ranks of F-Score and Nemenyi's critical difference (CD) all methods used in the experiments.}
\label{fig:CD-fscore}
\end{figure}

To evaluate the statistical significance of the F-Score results, we calculated the average Friedman ranks of MINOTAUR, HCM-GA, Clus-HMC and the performance-restricted baselines J48 (BR, CC and LP), SVM (BR, CC and LP), NB (BR, CC and LP), and k-NN (BR, CC and LP), totaling 15 methods (see Table~\ref{tab:fscores}). The ranks suggest there are statistical differences between the methods. Thus, we proceed with a post-hoc Nemenyi test to find which methods provide better results in a pairwise fashion. Figure~\ref{fig:CD-fscore} shows the critical diagram for comparing all methods in terms of F-Score. CD stands for the critical difference ($CD = 7.58$) at a confidence level of 95\%, and methods connected by a line do not present statistically significant differences. Since there is no line connecting SVM-LP, MINOTAUR, HMC-GA, Clus-HMC and J48-LP, we conclude that SVM-LP is significantly better than these methods in terms of F-Score. The other pairwise comparisons do not show significantly differences with statistical significance. 

It is important to emphasize that only SVM-LP was statistically superior to MINOTAUR. This is a very good result for our method given that SVM-LP is focused and tuned only to obtain good predictive performances. MINOTAUR obtained competitive results compared to SVM-LP with the advantage of generating interpretable models.
This can be confirmed by looking again at Table~\ref{tab:fscores}, where our proposal obtained very competitive results in some datasets such as CAL500, synthetic0 and synthetic1.


%

Since we have only three rule-based methods to be compared in terms of number of rules, we conducted a pairwise Wilcoxon test~\cite{wilcoxon1945individual} to assess the statistical significance. The null hypothesis states that the medians of the differences between the pair of algorithms do not differ. We run this test with a significance level of $10\%$, which is a standard value considering the fact that Wilcoxon's is quite conservative for indicating significant differences. The results are presented in Table~\ref{tab:wilcoxon_test}. We can see that MINOTAUR was statistically superior to both HMC-GA and Clus-HMC, thus generating more intepretable models.

\begin{table}[htbp]
\centering
\caption{Results of the Wilcoxon pairwise test regarding RuleCount. Underlined $p$-values indicate a refusal of the null hypothesis, i.e., a superiority of the row algorithm over the column one}
\label{tab:wilcoxon_test}
\begin{tabular}{lccc}
\toprule
&        MINOTAUR &        HMC-GA &      Clus-HMC  \\
\midrule                                                                                
MINOTAUR  & -- &  \underline{0.0781} &  \underline{0.0547}   \\
HMC-GA  &  0.0781 & -- & 0.5469    \\
Clus-HMC  &  0.0547 & 0.5469  &  --                \\
\bottomrule
\end{tabular}
\end{table}

\begin{figure*}[htbp]
   \center
   \subfigure[CAL500]{\includegraphics[scale=0.37]{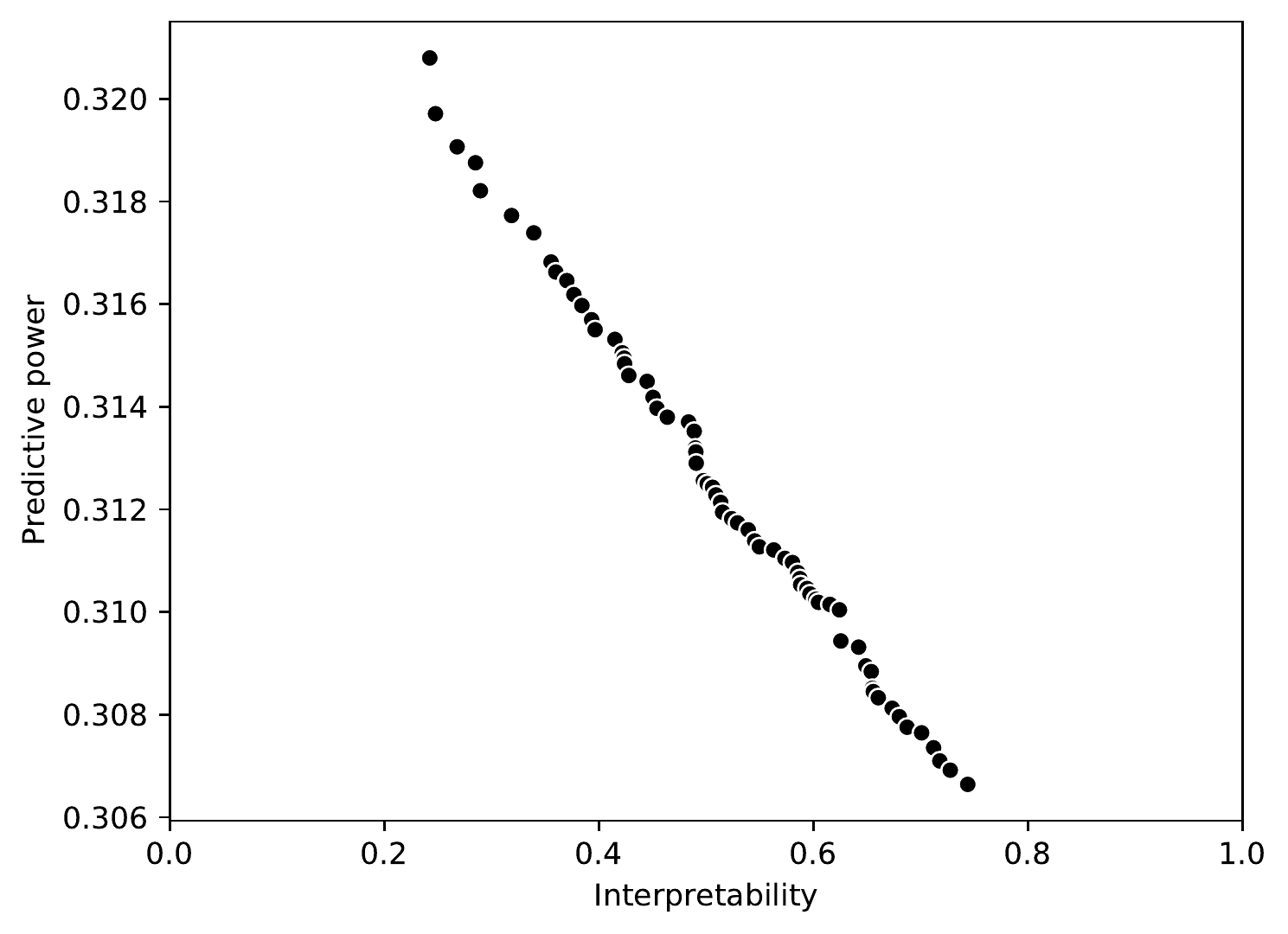}}
   \hspace{0.3em}
   \subfigure[emotions]{\includegraphics[scale=0.37]{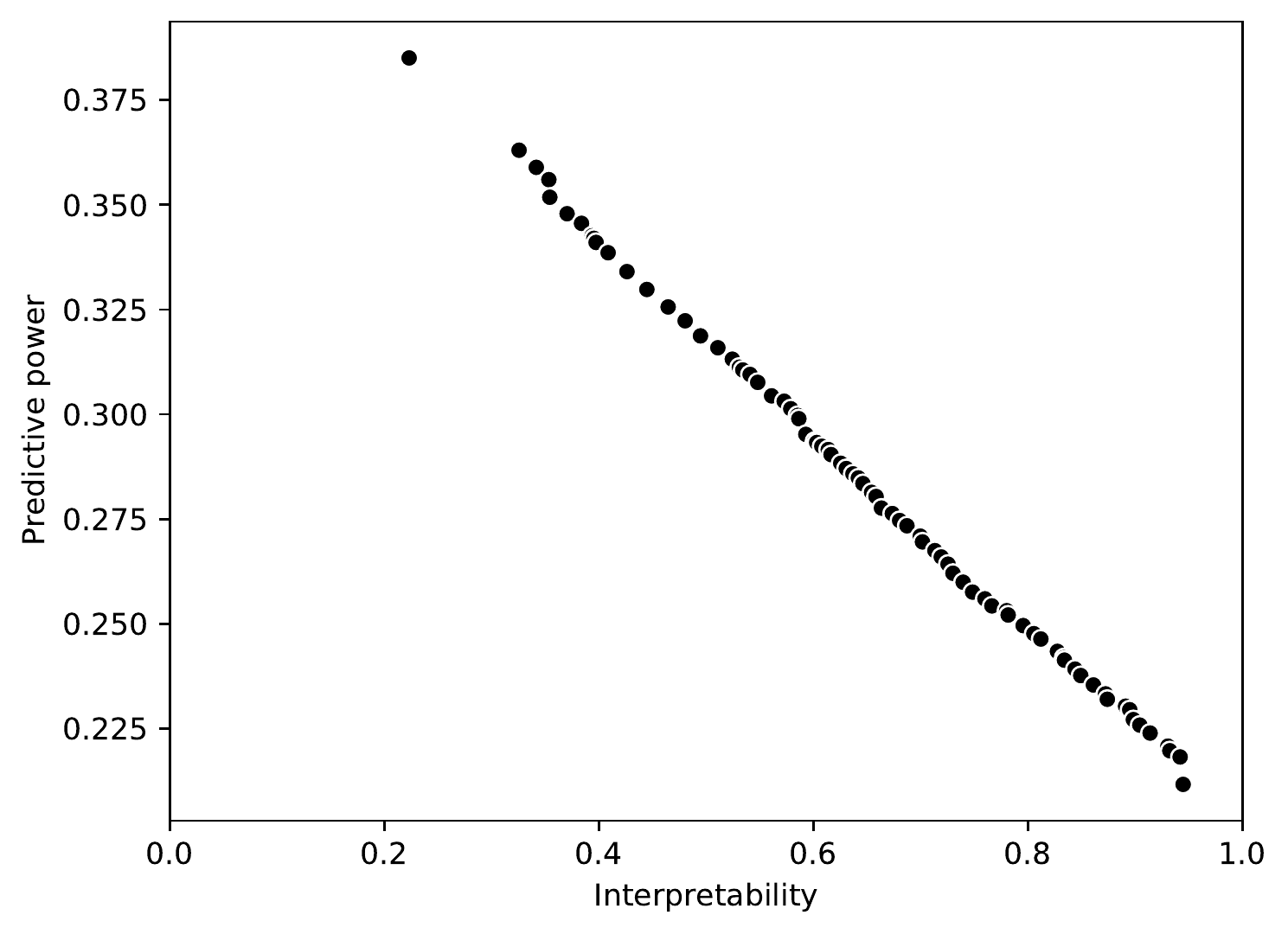}}
   \hspace{0.3em}
   \subfigure[scene]{\includegraphics[scale=0.37]{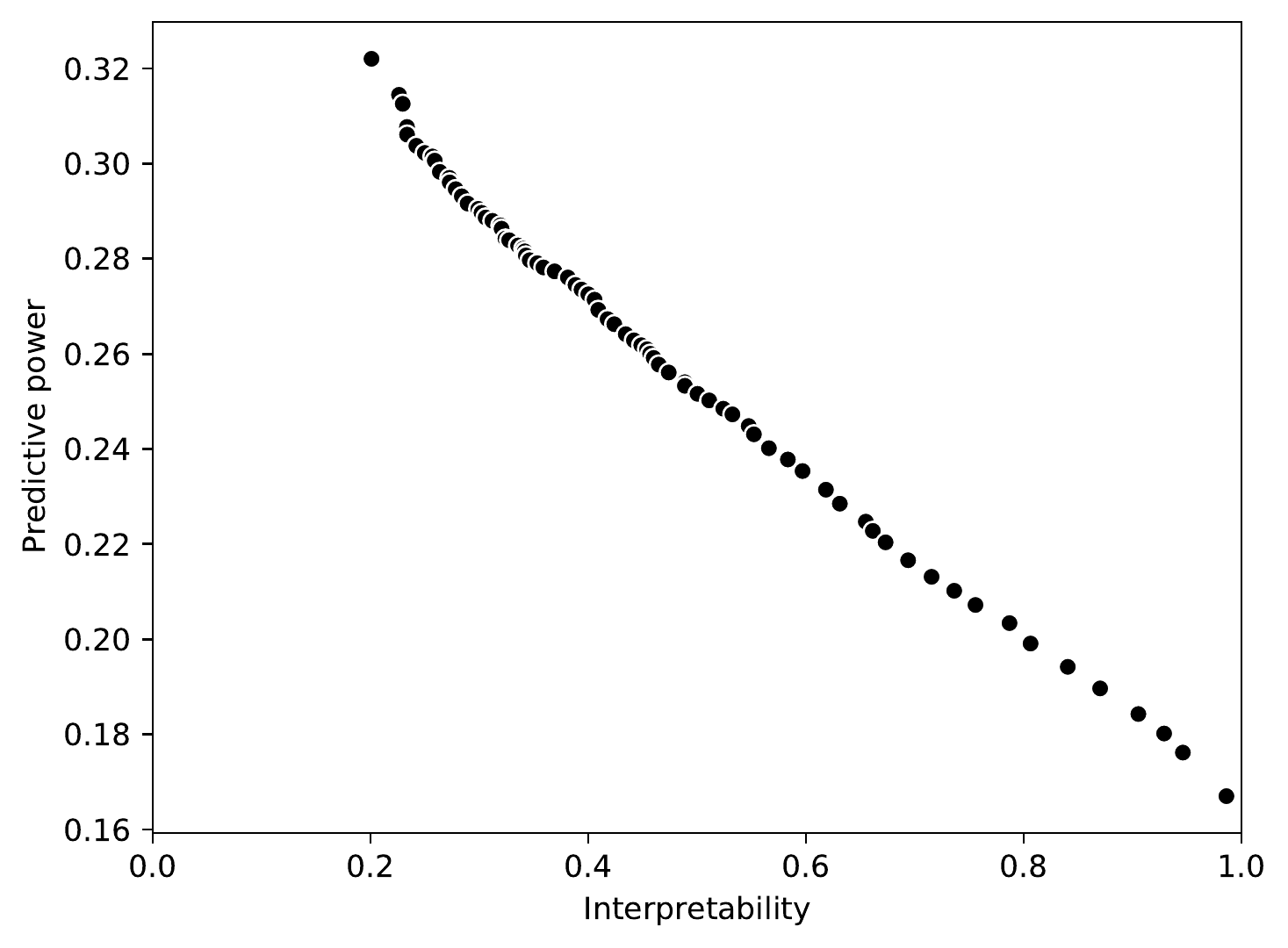}}
   \hspace{0.3em}
   \subfigure[synthetic0]{\includegraphics[scale=0.37]{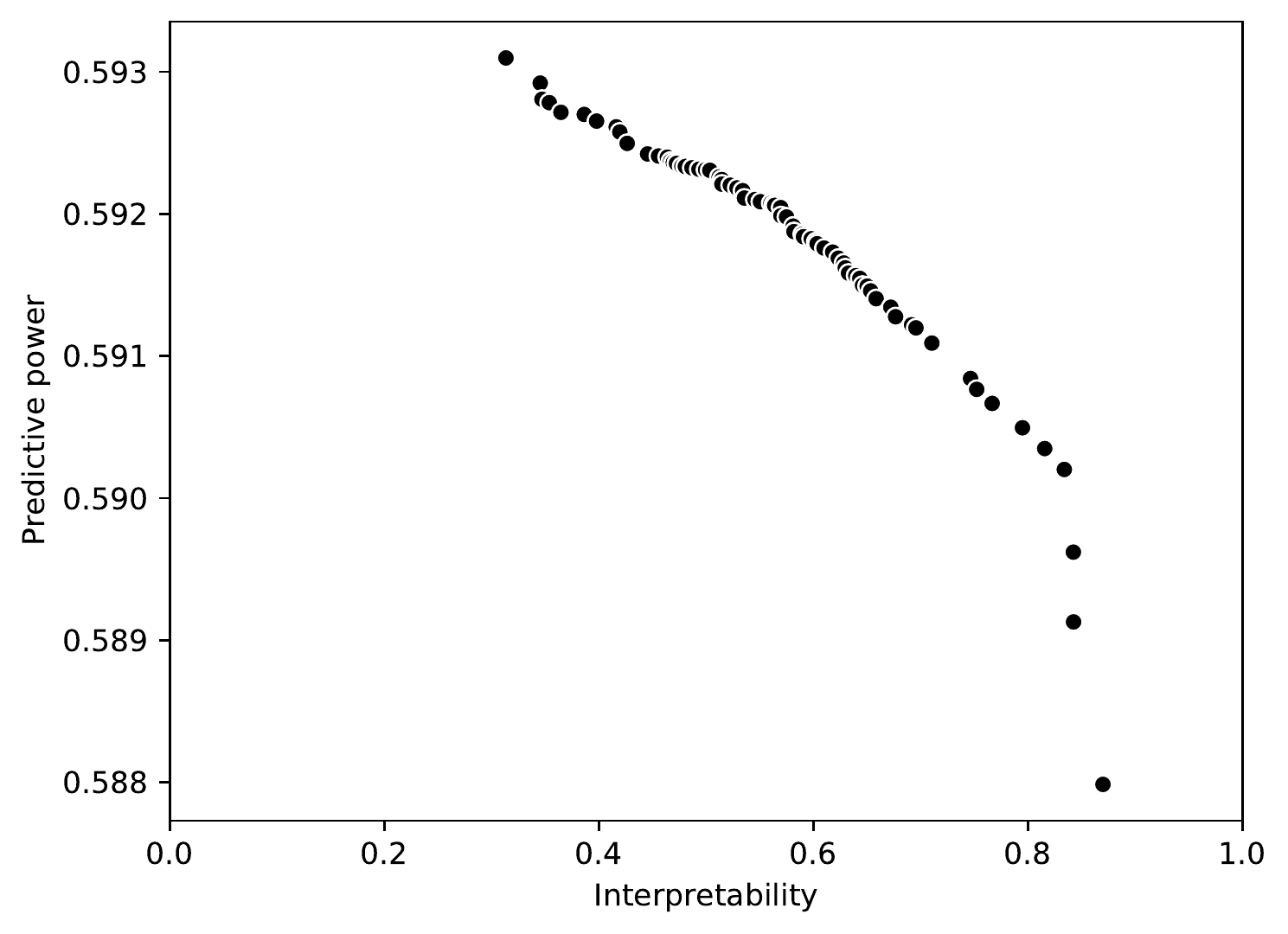}}
   \hspace{0.3em}
   \subfigure[synthetic1]{\includegraphics[scale=0.37]{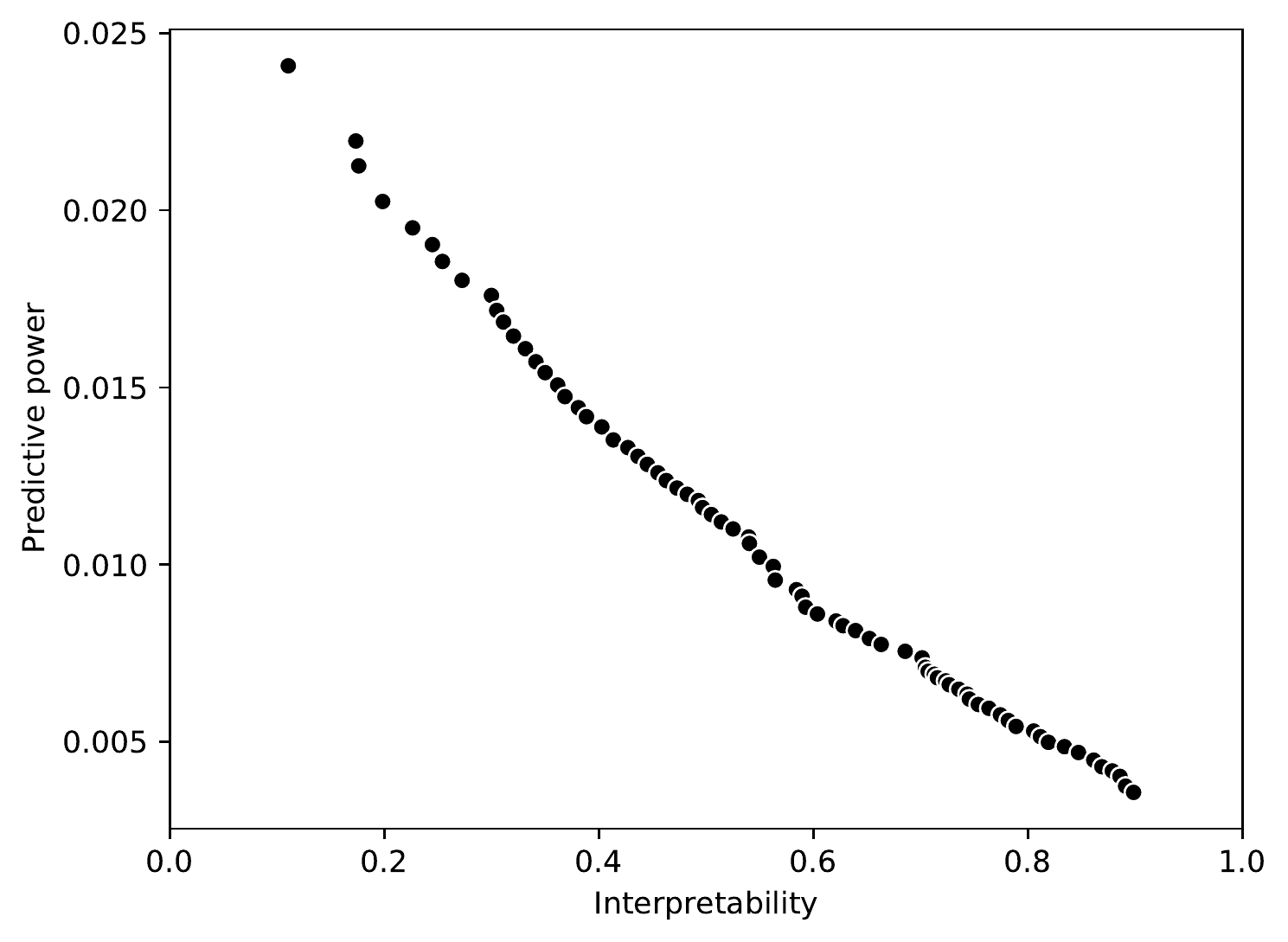}}
   \hspace{0.3em}
   \subfigure[synthetic2]{\includegraphics[scale=0.37]{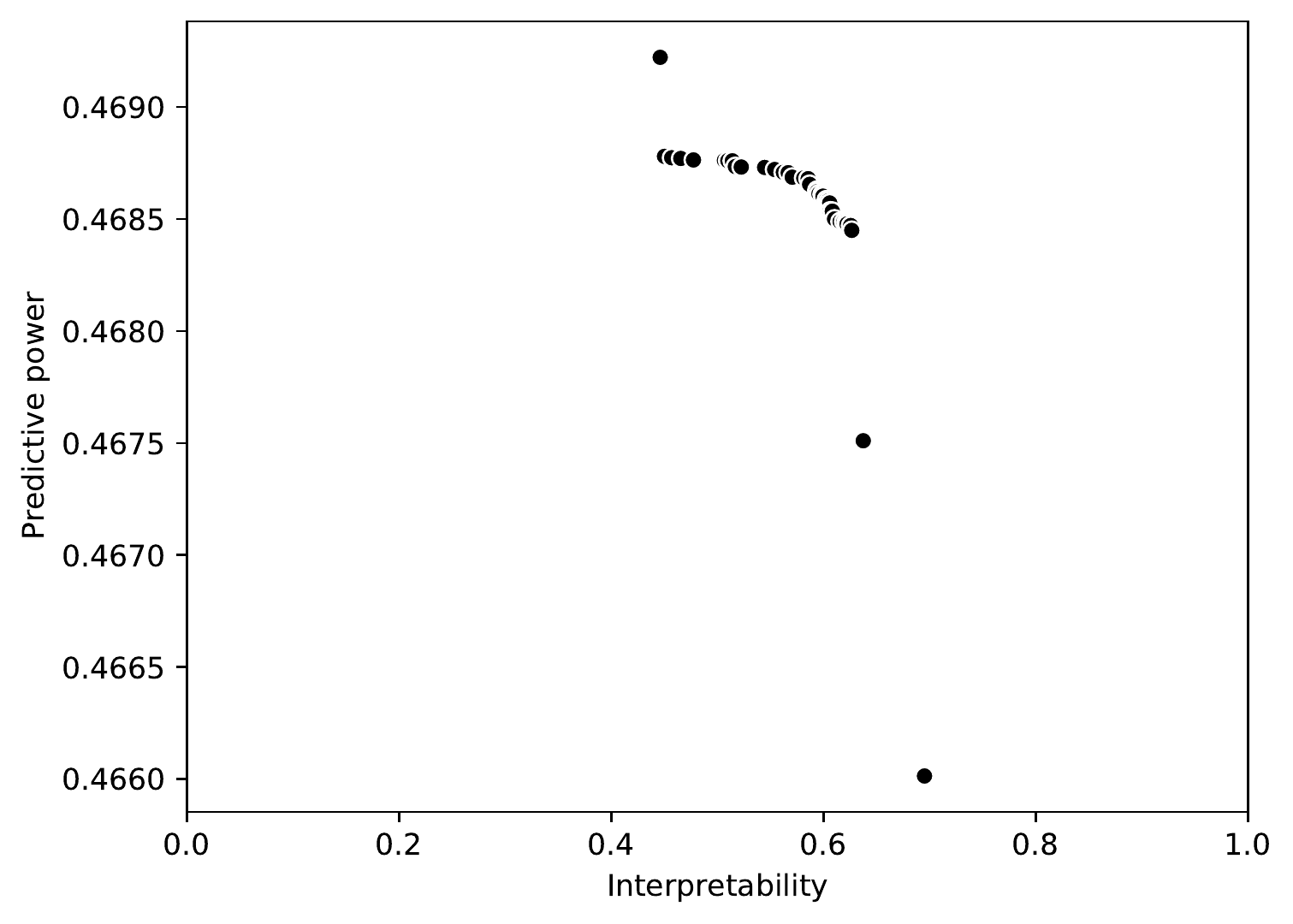}}
   \hspace{0.3em}
   \subfigure[synthetic3]{\includegraphics[scale=0.37]{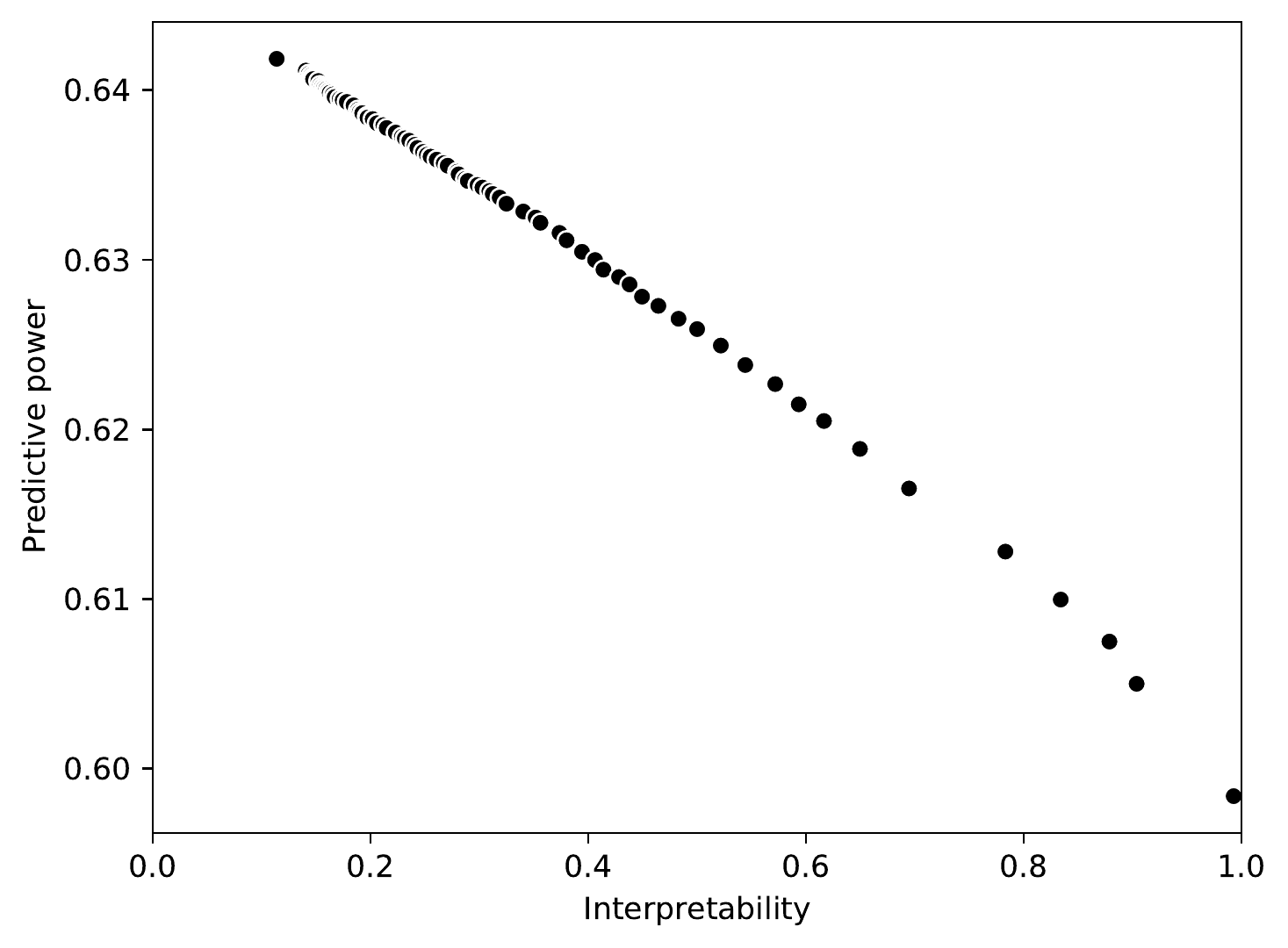}}
   \hspace{0.3em}
   \subfigure[yeast]{\includegraphics[scale=0.37]{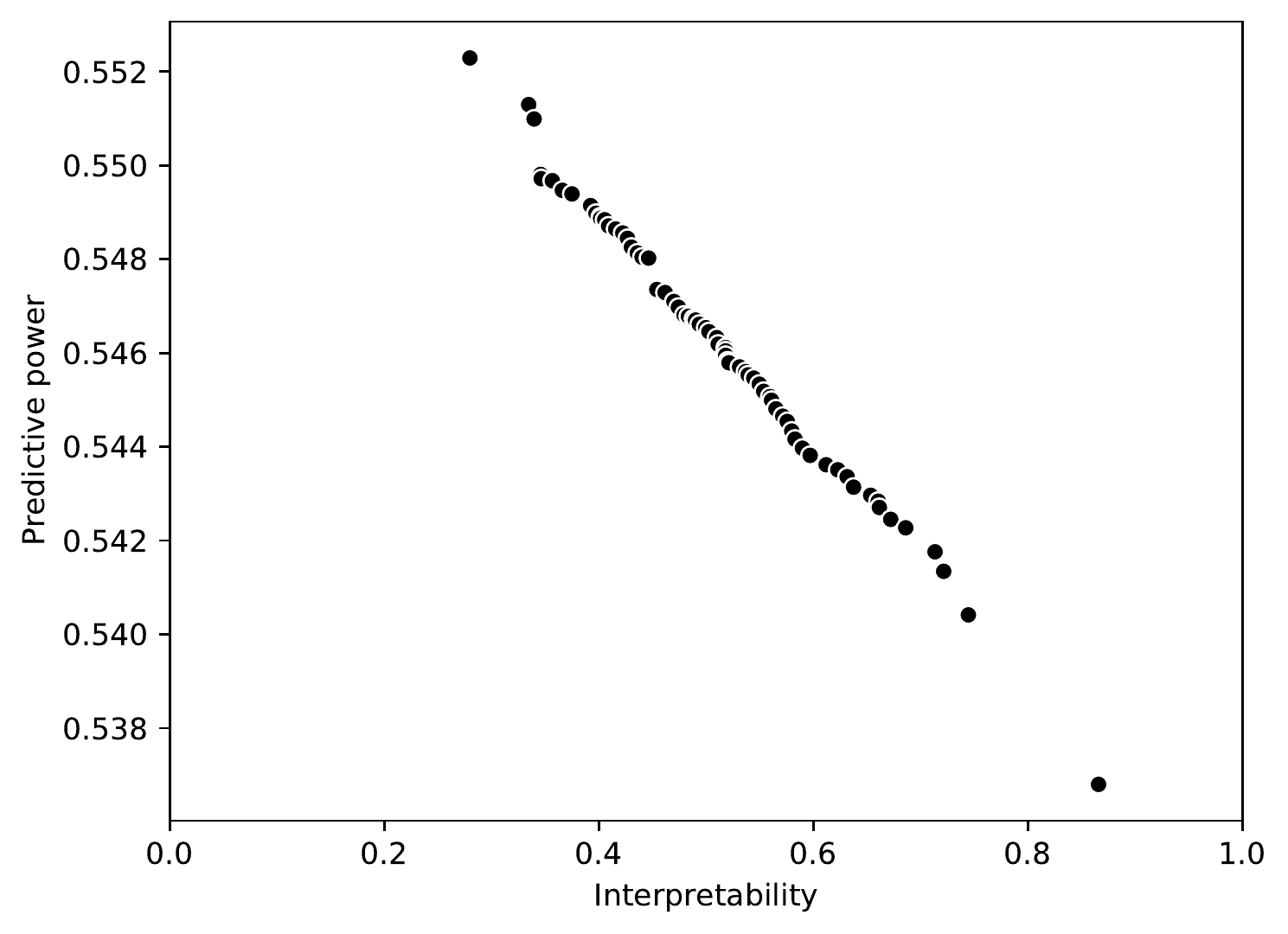}}
   \caption{Predictive power and interpretability compromises}
   \label{fig:compromises}
\end{figure*}

A particularly important aspect of MINOTAUR is that it generates a collection
of classification models, allowing its user to choose between models
with different compromises between predictive power and interpretability.
To facilitate the visualization of such aspect,
we generated the graphics shown in Figure~\ref{fig:compromises}.
In such graphs,
the horizontal axis (Interpretability)
is measured as the inverse of the model size (number of rules), that is,
the smallest (most interpretable) models are positioned in the rightmost part.
The vertical axis (Predictive Power) is measured as F-Score,
positioning the most powerful models in the upper part of the graphs. 

To generate each graph in Figure~\ref{fig:compromises},
we used the following procedure:
for each run (30), for each fold (10),
we select the fitness (a tuple with F-Score and model size)
of the individuals from the last generation.
Since the population size is 80 we obtain 300 matrices with dimension (80,2),
80 rows by 2 columns.
We then computed the average of such 300 matrices.
To constrain the model size to the $[0, 1]$ interval,
we substituted it for its inverse.
We then plotted the non-dominated rows of the matrix.

The graphs show that MINOTAUR is able to generate
individuals that offer different compromises between predictive power and interpretability, 
and that there is usually a negative correlation between such quantities.
We can see that in difficult datasets,
such as CAL500, the most powerful models may have two times more rules
than the most interpretable ones, but the difference in F-Score is smaller
than 0.02.
In relatively easier datasets, such as scene,
we can see that the largest models may contain five times more rules
than the smaller ones, with a F-Score two times higher.

\subsection{Reproducibility}
This section is dedicated to facilitate the reproduction of our experiments, which is, in our opinion, of paramount importance. The next paragraphs discuss the hyper-parameter values used in the proposed and baseline algorithms.

The hyper-parameters of the Mulan-based algorithms were set as their default values. Since Mulan\footnote{\url{http://mulan.sourceforge.net/}} extends Weka~\cite{Weka2009}, the J48, SVM, NB and k-NN hyper-parameters were the Weka default ones. Clus-HMC was executed with the default values from its original publication~\cite{vens2008decision}\footnote{\url{https://dtai.cs.kuleuven.be/clus/hmc-ens/}}. Regarding HMC-GA\footnote{\url{http://www.biomal.ufscar.br/resources.html}}, it is a conventional genetic algorithm. We used the hyper-parameter values recommended in Cerri et al., 2019~\cite{cerri2019inducing}: initial population size = 50, elitism number = 1, mutation rate = 40\%, crossover rate = 90\%, tournament size = 2, maximum number of generations = 50, maximum number of covered instances per rule = 300, minimum number of covered instances per rule = 5, and maximum uncovered instances (covered by the default rule) = 10. HMC-GA also has a hyper-parameter which generates a specific value for each dataset. It is a value $p$ which is used in the initialization phase (initial population) to define the probability of using a test in a rule. According to this probability $p$, HMC-GA generates initial rules with an average size (number of tests) $n = f \times p$, with $f$ the number of dataset features. Table~\ref{tab:hyperparameters} shows the average size $n$ of each rule in HMC-GA for $p = 0.3$, which was the value used in our experiments. The actual size of the rules can vary since during evolution HMC-GA applies a local search operator to decrease the size of the rules while keeping performance.

We ran MINOTAUR with the same population size (80),
maximum number of generations (200),
maximum number of failed mutation attempts per generation (2000)
and number of mutants to be generated per generation ($m = 40$)
for all datasets.
Only the hyper-parameter $t$ (see Section~\ref{sec:rule-creation})
was tuned specifically for each dataset.
We conducted experiments with $t \in \{2, 8, 16, 32, 64, 128, 256, 512, 1024, 2048, 4096\}$
and used the values (shown in Table~\ref{tab:hyperparameters})
that yielded the highest F-Score on the train~dataset.


\begin{table}[htbp]
\centering
\caption{Datasets Specific Hyper-Parameter values}
\begin{tabularx}{0.8\columnwidth}{Xll}
\toprule
Dataset     & MINOTAUR ($t$)& HMC-GA ($n$) \\
\midrule
CAL500 		& 64            & 20.4 \\
emotions 	& 128           & 21.6\\
scene 		& 512           & 88.2 \\
synthetic0 	& 4096          & 3 \\
synthetic1 	& 32            & 24 \\
synthetic2 	& 2048          & 9 \\
synthetic3 	& 1024          & 6.3 \\
yeast 		& 512           & 30.9 \\
\bottomrule
\end{tabularx}
\label{tab:hyperparameters}
\end{table}

The original datasets,
the scripts used to pre-process them
(which consisted of basic file format conversions, e.g.: from \textit{.csv} to \textit{.arff}),
and the generated folds are hosted on GitHub\footnote{\url{https://github.com/Mirandatz/arxiv.minotaur.datasets}}.
The scripts used to run MINOTAUR, parse its output and generated the graphics in Figure~\ref{fig:compromises} is also hosted on GitHub\footnote{\url{https://github.com/Mirandatz/arxiv.minotaur.experiments}}.


\section{Conclusion}
\label{sec:conclusion}
In this work,
we presented a multi-objective evolutionary algorithm
that generates classification models based on consistent sets of rules.
Our experiments indicated that the generated models
offer different compromises between predictive power and interpretability,
with the best models being competitive with state-of-the-art algorithms,
especially multi-label rule-based ones.

As future works we plan to improve some aspects of MINOTAUR, such as to implement different fittest selection mechanisms, and perform experiments with different mutation weights. Also, the current implementation of MINOTAUR has a few limitations: it only handles continuous features, and has no default values for its hyper-parameters, nor a built-in heuristic to generate such values. 
We plan to improve the method regarding these limitations.



\bibliographystyle{IEEEtran}
\bibliography{0-main} 

\end{document}